\documentclass[a4paper]{article}

\usepackage{INTERSPEECH2022}
\usepackage{float}
\usepackage{pgfplots}
\pgfplotsset{compat=1.18}
\pgfplotsset{width=7cm,compat=1.3}
\usepackage{adjustbox}
\usepackage{multirow}
\usepackage{threeparttable}
\usepackage{caption}
\usepackage[hidelinks]{hyperref}
\usepackage{url}
\usetikzlibrary{patterns}

\title{Speaker-Adapted End-to-End\\Visual Speech Recognition for Continuous Spanish}
\name{David Gimeno-Gómez , Carlos-D. Martínez-Hinarejos}
\address{
  Pattern Recognition and Human Language Technologies Research Center,\\
  Universitat Politècnica de València, Camino de Vera, s/n, 46022, València, Spain
  }
\email{dagigo1@dsic.upv.es, cmartine@dsic.upv.es}

\begin{document}

\maketitle
\begin{abstract}
Different studies have shown the importance of visual cues throughout the speech perception process. In fact, the development of audiovisual approaches has led to advances in the field of speech technologies. However, although noticeable results have recently been achieved, visual speech recognition remains an open research problem. It is a task in which, by dispensing with the auditory sense, challenges such as visual ambiguities and the complexity of modeling silence must be faced. Nonetheless, some of these challenges can be alleviated when the problem is approached from a speaker-dependent perspective. Thus, this paper studies, using the Spanish LIP-RTVE database, how the estimation of specialized end-to-end systems for a specific person could affect the quality of speech recognition. First, different adaptation strategies based on the fine-tuning technique were proposed. Then, a pre-trained CTC/Attention architecture was used as a baseline throughout our experiments. Our findings showed that a two-step fine-tuning process, where the VSR system is first adapted to the task domain, provided significant improvements when the speaker adaptation was addressed. Furthermore, results comparable to the current state of the art were reached even when only a limited amount of data was available.
\end{abstract}

\section{Introduction}
Influenced by studies which have demonstrated the relevance of visual cues throughout the speech perception process \cite{mcgurk1976hearing} , different advances have been achieved in the field of Speech Technologies. The robustness of automatic speech recognition systems, specially in adverse conditions where the acoustic signal was damaged or corrupted \cite{juang1991adverse}, has been enhanced by the design of audio-visual approaches \cite{afouras2018deep,ma2022visual}. Moreover, these studies have encouraged the development of systems capable of interpreting speech by reading only the lips of the speaker. In fact, this challenging task, known as Visual Speech Recognition (VSR), has been a focus of interest during the last decades \cite{fernandez2018survey}. 

Nowadays, remarkable results have been achieved in the field of VSR \cite{ma2022visual,afouras2021sub}. Both the design of end-to-end architectures and the availability of large-scale databases have been fundamental pillars of recent advances in the field \cite{fernandez2018survey}. However, the VSR task remains an open research problem where, by dispensing with the auditory sense, challenges such as visual ambiguities and the complexity of modeling silence must be faced. In fact, Duchnowski et al. \cite{duchnowski2000development} maintained that only 30\% of speech information is visible. Furthermore, it has been proven that each person produces speech in a unique way \cite{leung04_interspeech}. This fact supports the idea that visual speech features are considered as highly sensitive to the identity of the speaker \cite{cox2008challenge}, which poses an additional challenge when estimating speaker-independent VSR systems.

Nonetheless, these challenges can be alleviated when the VSR task is approached from a speaker-dependent perspective \cite{thangthai2017,assael2016lipnet}. As detailed in Section \ref{relatedwork}, there is a wide range of works which have studied the speaker adaptation of end-to-end systems in the field of Acoustic Speech Recognition (ASR) \cite{watanabe2018adaptation,delcroix18_interspeech,weninger19b_interspeech,ctc2018adaption}. On the contrary, few works in this regard have been addressed in VSR \cite{kandala19_interspeech,cogans2020adriana}. Albeit this speaker-dependent approach means facing a less demanding task, it should not be forgotten that speaker-adapted VSR systems could be helpful, in a non-invasive and inconspicuous way, for people who suffer from communication difficulties \cite{ssi2020review,kenji2020enhancement}.

On the other hand, addressing VSR for languages other than English is recently receiving an increasing interest \cite{fernandez2017towards,gimenogomez21_iberspeech,ma2022visual}. However, in these cases, the lack of audio-visual resources that it entails \cite{lrec2022liprtve,fernandez2018survey} must be taken into account.

\textbf{Contributions:} in this paper, a study related to VSR for continuous Spanish is presented. The LIP-RTVE database \cite{lrec2022liprtve} was used throughout all our experiments. All the defined VSR systems were based on a hybrid CTC/Attention architecture \cite{ma2022visual}, which was pre-trained with hundreds of hours of data. Then, we studied how the estimation of specialized end-to-end systems for a specific person affects the quality of speech recognition. Hence, different adaptation strategies based on the fine-tuning technique were proposed. Our findings showed that significant improvements could be obtained when addressing the speaker adaptation through a two-step fine-tuning process, where the VSR system is first adapted to the task domain. Furthermore, even when only a limited amount of data was available, results comparable to the current state of the art were reached. 

\section{Related Work} \label{relatedwork}

This section offers a brief overview of the VSR task in the literature, as well as of the different works that have addressed the speaker adaption problem for speech recognizers based on end-to-end architectures. Finally, research on VSR for the Spanish language to date is considered.

\textbf{Visual Speech Recognition:} influenced by the evolution of systems focused on acoustic speech recognition, different approaches were considered in the field \cite{fernandez2018survey}. Nowadays, the current state of the art in VSR \cite{ma2022visual,afouras2021sub} has shown remarkable advances, achieving around a 70\% of word recognition rate on the challenging LRS3-TED database \cite{afouras2018lrs3}. This has been possible not only to the availability of large-scale databases, but also to the design of appropriate architectures and the definition of adequate optimisation methods \cite{ma2022visual,fernandez2018survey}.

\textbf{Speaker Adaptation of End-to-End Architectures:} although speaker adaptation has been widely studied with traditional paradigms \cite{gales2008application,ivectors2013}, in this section we only consider, due to the nature of our VSR system, those works that addressed the problem with end-to-end architectures. A simple retraining-based adaptation was adopted in \cite{watanabe2018adaptation} to fine-tune an Attention-based system. Besides, influenced by research done on conventional ASR systems \cite{ivectors2013}, a hybrid CTC/Attention model was adapted by the incorporation of speaker identity vectors \cite{delcroix18_interspeech}. On the other hand, some works \cite{weninger19b_interspeech,ctc2018adaption} proposed the use of more sophisticated techniques such as the Kullback-Leibler divergence and Linear Hidden Networks to adapt a pre-trained speaker-independent system.

However, most of these studies were conducted in the ASR domain. Albeit this research describes approaches that could be adopted to any speech modality, it is noteworthy that few works have focused specifically on speaker adaptation for VSR systems. Kandala et al. \cite{kandala19_interspeech} defined an architecture based on the Connectionist Temporal Classification (CTC) paradigm \cite{graves2006ctc} where, once visual speech features were computed, a speaker-specific identity vector was integrated as an additional input to the decoder. Moreover, Fernandez-Lopez et al. \cite{cogans2020adriana} approached the problem indirectly, since it was studied how to adapt the visual front-end of an audio-visual recognition system. Thus, the authors proposed an unsupervised method that allowed an audiovisual system to be adapted when only the visual channel was available. Nevertheless, unlike our research, these works did not addressed natural continuous VSR because their experiments were evaluated on databases that were recorded in controlled settings.

\textbf{Spanish Visual Speech Recognition:} although they still suffer from a lack of audiovisual resources, other languages besides English are beginning to be considered in the field \cite{zadeh2020moseas,ma2022visual}. This is the case of the Spanish language which, despite the fact that an evaluation benchmark has not yet been specified, has been the object of study on multiple occasions. Fernandez-Lopez \cite{adriana2022alr} explored diverse approaches over the VLRF corpus \cite{fernandez2017towards}, achieving around 30\% accuracy at word level in the best setting of their experiments. Besides, Ma et al. \cite{ma2022visual} designed a hybrid CTC/Attention architecture that, after being pre-trained with large-scale English corpora, was fine-tuned with the Spanish CMU-MOSEAS database \cite{zadeh2020moseas}, achieving a word error rate of approximately 45\%.

Regarding our previous work, we carried out several experiments where different visual speech representations were studied \cite{gimenogomez21_iberspeech}. Subsequently, we compiled the challenging LIP-RTVE database \cite{lrec2022liprtve}, an audiovisual corpus primarily conceived to deal with the Spanish VSR task and whose details are described in Section \ref{database}. Additionally, baseline results were reported in \cite{lrec2022liprtve}, using a traditional paradigm based on Hidden Markov Models \cite{gales2008application}. More specifically, we reached around 80\% of word error rate for the speaker-dependent partition, but we were not able to obtain acceptable results (about 95\% error rate) for the speaker-independent scenario. Finally, parallel to this study, we have improved the aforementioned performances in the order of roughly 40\% for both scenarios, using the pre-trained CTC/Attention architecture publicly released by \cite{ma2022visual}.

\section{Proposed Study} \label{proposal}

The purpose of our research is to study the feasibility of developing speaker-adapted systems for the VSR task. In our case, we are going to consider the speaker dependent scenario defined for the LIP-RTVE database. Thus, we analyze how the estimation of specialized end-to-end systems for a specific person could affect the quality of speech recognition in this scenario. Hence, three different training strategies were proposed:

\begin{itemize}

    \item \textbf{Multi-Speaker Training (MST):} the VSR system is re-estimated using the whole speaker-dependent training data; it can be seen as a task adaptation.
    \item \textbf{Speaker-Adapted Training (SAT):} in this strategy only data corresponding to a specific speaker is considered when fine tuning the VSR system.
    \item \textbf{Two-Step Speaker-Adapted Training (TS-SAT):} as its name suggests, this method consists of two fine-tuning steps. First, following the MST strategy, training data of the whole set of speakers is used to re-train the VSR system and achieve task adaptation. Afterwards, the system is fine-tuned to a specific speaker using her/his corresponding data.
    
\end{itemize}

In this way, by using a speaker-dependent database, we are able to study how a VSR system can generalize common patterns from different speakers or, on the contrary, evaluate to what extent it is capable of adapting to a specific speaker. 

Furthermore, in order to estimate all these VSR systems, we used a simple retraining-based method, also known as fine-tuning, whose details are specified in Subsection \ref{training}. This decision was mainly influenced by our GPU memory constraints described in Subsection \ref{implementation}.



\section{The LIP-RTVE Database} \label{database}

One of the main reasons why we have chosen the LIP-RTVE database\footnote{\url{https://github.com/david-gimeno/LIP-RTVE}} is because it offers, for the Spanish language, a suitable support to estimate VSR systems against realistic scenarios. It is a challenging database compiled from TV newscast programmes that were recorded at 25 fps with resolution of 480$\times$270 pixels. No type of restriction was considered in data collection, being able to find the so-called spontaneous speech phenomena, as well as different lighting conditions or head movements. The corpus is composed of 323 speakers, providing 13 hours of data with a vocabulary size of 9308 words.

\section{Model Architecture} \label{architecture}

The model architecture employed in our research is based on the work developed in \cite{ma2022visual}. The following modules are distinguished:

\begin{itemize}
    \item \textbf{Visual Front-end:} it consists of a 2D ResNet-18 \cite{he2016resnet} whose first layer, in order to deal with data temporal relationships, has been replaced by a 3D convolutional layer.
    
    \item \textbf{Conformer Encoder:} a 12-layer Conformer \cite{gulati20_interspeech} block is defined to capture global and local speech interactions from the previous latent visual representation.
    
    \item \textbf{Hybrid CTC/Attention Decoder:}  it is composed of a 6-layer Transformer \cite{vaswani2017attention} block and a fully connected layer as the CTC-based decoding branch \cite{graves2006ctc}.
    
    \item \textbf{Language Model:} a character-level language model (LM), composed of 6 Transformer \cite{vaswani2017attention} layers, is integrated during the decoding process. 
    
\end{itemize}

The entire model is estimated according to a loss function that combines both the CTC and the Attention paradigm, an approach that has led to advances in speech processing \cite{watanabe2017ctcattention,petridis2018ctc-attn}.

\section{Experimental Settings}

\subsection{Data Sets}

The LIP-RTVE database defines a partition both for a speaker-dependent and speaker-independent scenario \cite{lrec2022liprtve}, each with its respective training, development and test sets. Due to the nature of the proposed study, we decided to use the speaker-dependent partition throughout our experiments. Nonetheless, in order to estimate our speaker-adapted VSR systems and properly interpret the obtained results, the 20 most talkative speakers were selected. Thus, albeit the MST-based VSR system was trained using the entire speaker-dependent partition (roughly 9 hours of data), each SAT-based model used only the data of its corresponding speaker (about 15 minutes for training and 3 minutes for development, in average terms per speaker). 

\subsection{Region of Interest Extraction}

By using state-of-the-art open source resources\footnote{\url{https://github.com/hhj1897/face_alignment}}\textsuperscript{,}\footnote{\url{https://github.com/hhj1897/face_detection}} \cite{deng2020retina,bulat2017facealign}, gray-scale bounding boxes centered on the speaker's mouth were cropped as images of 96$\times$96 pixels. The regions not only cover the mouth, but the complete jaw and the cheeks of the speaker, which has shown benefits when addressing VSR \cite{zhang2020rois}.

\subsection{Data Augmentation}

The data augmentation process was defined according to the guidelines described in \cite{ma2022visual}. First, once the ROIs were normalized with respect to the overall mean and variance of the training set, a random cropping of 88$\times$88 pixels was applied. Then, additional techniques such as horizontal flipping and time masking were considered. 

\subsection{Pre-Training} \label{pre-training}

The parameters publicly released by Ma et al. \cite{ma2022visual} for the Spanish CMU-MOSEAS database \cite{zadeh2020moseas} were used to pre-train both the VSR system and the LM described in Section \ref{architecture}. Concretely, the VSR system was pre-trained on over a thousand hours of data collected from multiple large-scale English corpora. Subsequently, by using the Spanish part of the CMU-MOSEAS and the Multilingual-TEDx database \cite{salesky21_interspeech}, the system was fine-tuned to the Spanish language. Regarding the LM, it was estimated from text collected from different databases until gathering more than 200 million characters.

\subsection{Training \& Decoding Setup} \label{training}
Our learning method consists of a simple retraining-based adaptation. Thus, the pre-trained VSR system was fine-tuned during 5 epochs, using the AdamW optimiser\footnote{\url{https://pytorch.org/docs/stable/generated/torch.optim.AdamW.html}} and a linear One Cycle Learning Rate Scheduler\footnote{\url{https://pytorch.org/docs/stable/generated/torch.optim.lr_scheduler.OneCycleLR.html}}. The learning rate was set to 5$\times$10\textsuperscript{-4} and the batch size, due to the memory constraints described in Subsection \ref{implementation}, had to consider only one sample. The rest of the hyperparameters related with the training process kept the default configuration, as detailed in \cite{ma2022visual}.

In the same way, most of the decoding hyperparameters kept their default settings \cite{ma2022visual}. Nonetheless, due to our memory constraints, the beam size value was reduced to 10.



\subsection{Methodology} \label{method}

\begin{figure*}
    \vspace{-0.30cm}
    \centering
    \resizebox{0.9\textwidth}{!}{
    \begin{tikzpicture}[ 
       declare function={
        barW=6pt; 
        barShift=barW/2; 
      }
    ]
    \begin{axis}[
        ybar,
        bar width=barW, 
        bar shift=-barShift*2, 
        symbolic x coords={spkr00, spkr01, spkr02, spkr03, spkr07, spkr09,
                           spkr10, spkr11, spkr12, spkr13, spkr17, spkr19,
                           spkr22, spkr24, spkr25, spkr26, spkr33, spkr42,
                           spkr45, spkr51},
        axis y line*=left,
        axis x line=none,
        ymin=0, ymax=60,
        minor y tick num=1,
        ylabel=\%WER,
        enlarge x limits=0.04,
        x=1cm,
        xtick={spkr00, spkr01, spkr02, spkr03, spkr07, spkr09,
                           spkr10, spkr11, spkr12, spkr13, spkr17, spkr19,
                           spkr22, spkr24, spkr25, spkr26, spkr33, spkr42,
                           spkr45, spkr51}
     ]
        \addplot[mark=*,
                 mark options={xshift=-barShift*2}, 
                 draw=black,
                 fill=black!60,
                 error bars/.cd,
                   y dir=both,
                   y explicit,
                   error bar style={line width=1pt,solid, black}
        ] coordinates {
            (spkr00, 39.5) +- (2.1, 2.1)
            (spkr01, 45.2) +- (4.5, 4.5)
            (spkr02, 27.4) +- (2.6, 2.6)
            (spkr03, 42.1) +- (4.7, 4.7)
            (spkr07, 45.1) +- (8.1, 8.1)
            (spkr09, 46.4) +- (7.4, 7.4)
            (spkr10, 38.8) +- (6.4, 6.4)
            (spkr11, 35.2) +- (8.5, 8.5)
            (spkr12, 20.3) +- (4.8, 4.8)
            (spkr13, 35.9) +- (6.3, 6.3)
            (spkr17, 35.7) +- (7.3, 7.3)
            (spkr19, 16.7) +- (8.7, 8.7)
            (spkr22, 32.9) +- (6.1, 6.1)
            (spkr24, 39.3) +- (10.5, 10.5)
            (spkr25, 19.0) +- (7.2, 7.2)
            (spkr26, 30.1) +- (7.0, 7.0)
            (spkr33, 48.5) +- (6.2, 6.2)
            (spkr42, 31.3) +- (10.4, 10.4)
            (spkr45, 19.2) +- (10.9, 10.9)
            (spkr51, 40.2) +- (6.9, 6.9)
        }; \label{mst}
    \end{axis}
    \begin{axis}[
        ybar,
        bar width=barW,
        symbolic x coords={spkr00, spkr01, spkr02, spkr03, spkr07, spkr09,
                           spkr10, spkr11, spkr12, spkr13, spkr17, spkr19,
                           spkr22, spkr24, spkr25, spkr26, spkr33, spkr42,
                           spkr45, spkr51},
        axis y line*=right,
        ymin=0, ymax=60,
        minor y tick num=1,
        ylabel=\%WER,
        enlarge x limits=0.04,
        x=1cm,
        x tick label style={rotate=60,anchor=east},
        xtick={spkr00, spkr01, spkr02, spkr03, spkr07, spkr09,
                           spkr10, spkr11, spkr12, spkr13, spkr17, spkr19,
                           spkr22, spkr24, spkr25, spkr26, spkr33, spkr42,
                           spkr45, spkr51}
    ]
    \addplot[mark=*,
             fill=black!0,
             error bars/.cd,
               y dir=both,
               y explicit,
               error bar style={line width=1pt,solid, black}
    ] coordinates {
            (spkr00, 36.5) +- (2.1, 2.1)
            (spkr01, 38.1) +- (5.3, 5.3)
            (spkr02, 22.6) +- (3.0, 3.0)
            (spkr03, 31.0) +- (4.7, 4.7)
            (spkr07, 27.8) +- (7.7, 7.7)
            (spkr09, 36.7) +- (9.5, 9.5)
            (spkr10, 32.8) +- (8.4, 8.4)
            (spkr11, 16.3) +- (8.5, 8.5)
            (spkr12, 18.2) +- (5.1, 5.1)
            (spkr13, 12.8) +- (8.5, 8.5)
            (spkr17, 18.2) +- (7.7, 7.7)
            (spkr19, 8.6) +- (6.2, 6.2)
            (spkr22, 22.5) +- (6.7, 6.7)
            (spkr24, 31.3) +- (10.4, 10.4)
            (spkr25, 12.5) +- (6.7, 6.7)
            (spkr26, 18.2) +- (6.0, 6.0)
            (spkr33, 42.8) +- (7.3, 7.3)
            (spkr42, 28.2) +- (11.3, 11.3)
            (spkr45, 15.2) +- (7.5, 7.5)
            (spkr51, 37.8) +- (9.0, 9.0)
        }; \label{sat};
    \end{axis}
    \begin{axis}[
        ybar,
        bar width=barW,
        bar shift=barShift*2, 
        legend style={at={(0.5,0.84)},anchor=south, legend columns=-1,text width=1.10cm,text height=1.5ex},
        symbolic x coords={spkr00, spkr01, spkr02, spkr03, spkr07, spkr09,
                           spkr10, spkr11, spkr12, spkr13, spkr17, spkr19,
                           spkr22, spkr24, spkr25, spkr26, spkr33, spkr42,
                           spkr45, spkr51},
        axis y line*=right,
        yticklabels={,,},
        axis line style={opacity=0},
        ymin=0, ymax=60,
        minor y tick num=1,
        ylabel=,
        enlarge x limits=0.04,
        x=1cm,
        x tick label style={rotate=60,anchor=east},
        xtick={spkr00, spkr01, spkr02, spkr03, spkr07, spkr09,
                           spkr10, spkr11, spkr12, spkr13, spkr17, spkr19,
                           spkr22, spkr24, spkr25, spkr26, spkr33, spkr42,
                           spkr45, spkr51}
    ]
        \addlegendimage{/pgfplots/refstyle=mst}\addlegendentry{MST}
        \addlegendimage{/pgfplots/refstyle=sat}\addlegendentry{SAT}
        \addplot[mark=*,
                 postaction={pattern=north east lines},
                 mark options={xshift=barShift*2}, 
                 fill=black!0,
                 error bars/.cd,
                   y dir=both,
                   y explicit,
                   error bar style={line width=1pt,solid,black}
        ] coordinates {
                (spkr00, 32.5) +- (2.4, 2.4)
                (spkr01, 28.9) +- (5.5, 5.5)
                (spkr02, 19.2) +- (3.1, 3.1)
                (spkr03, 29.2) +- (4.9, 4.9)
                (spkr07, 22.2) +- (7.3, 7.3)
                (spkr09, 30.3) +- (8.7, 8.7)
                (spkr10, 29.2) +- (7.0, 7.0)
                (spkr11, 13.7) +- (8.2, 8.2)
                (spkr12, 13.6) +- (5.0, 5.0)
                (spkr13, 6.6) +- (5.0, 5.0)
                (spkr17, 13.5) +- (5.2, 5.2)
                (spkr19, 5.7) +- (4.3, 4.3)
                (spkr22, 17.9) +- (6.7, 6.7)
                (spkr24, 27.4) +- (10.3, 10.3)
                (spkr25, 8.7) +- (5.7, 5.7)
                (spkr26, 13.2) +- (4.7, 4.7)
                (spkr33, 38.6) +- (7.2, 7.2)
                (spkr42, 26.7) +- (12.4, 12.4)
                (spkr45, 12.0) +- (7.3, 7.3)
                (spkr51, 35.0) +- (9.0, 9.0)
            }; \label{ts-sat}; \addlegendentry{TS-SAT}
    \end{axis}
    \end{tikzpicture}}
    \vspace{-0.15cm}
    \caption{Comparison of the proposed adaptation strategies. System performance (WER) with 95\% confidence intervals is reported for each speaker considered in the study. Only experiments that used the training data set to estimate the VSR systems are considered.} \label{persp}
    \label{fig:comparision}
    \vspace{-0.45cm}
\end{figure*}
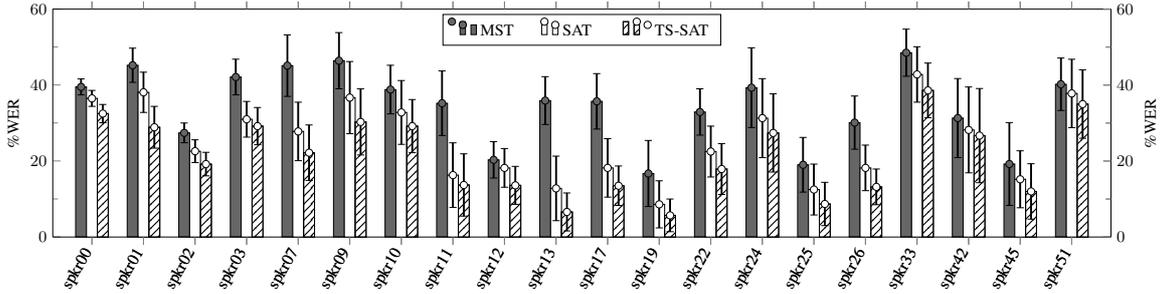

Before reporting and discussing our results, different aspects on how we carried out our experiments must be clarified:

\begin{itemize}

    \item Both the MST- and the SAT-based systems were initially pre-trained with the parameters released by Ma et al. \cite{ma2022visual}, as described in Subsection \ref{pre-training}.

    \item The MST-based system was fine-tuned using the entire speaker-dependent partition of the LIP-RTVE database.
    
    \item A SAT-based system was independently estimated for each speaker considered in our study, i.e., 20 SAT-based systems were defined.
    
    \item Regarding the TS-SAT strategy, the previously estimated MST-based system was used as a starting point, which could be considered as an adaptation of the model to the task. Afterwards, we followed the same fine-tuning scheme described with the SAT strategy to obtain a TS-SAT-based system for each speaker.
    
    \item Experiments were conducted using either the training or the development set for adapting, as reflected in Table \ref{tab:overall}. However, it should be noted that TS-SAT-based systems were always based on the MST-based system estimated with the training set, while in the second step training or validation were used to fine-tune depending on the experiment.
    
    \item All these VSR systems, as Figure \ref{fig:comparision} shows, were evaluated on the test set corresponding to each of the speakers selected in our study. The MST-based system was the same regardless the evaluated speaker. Conversely, for the rest of strategies, the corresponding speaker-adapted system was used in each case.
    
    \item The LM used in all the tests was the one described in Subsection 6.4.
    
\end{itemize}



\vspace{-0.30cm}
\subsection{Evaluation Metric}

All the results reported in our experiments were evaluated by the well-known Word Error Rate (WER) with 95\% confidence intervals obtained by the bootstrap method as described in \cite{bisani2004bootstrap}.
\vspace{-0.50cm}
\subsection{Implementation Details} \label{implementation}

The VSR system was implemented using the PyTorch\footnote{\url{https://pytorch.org/}} back-end of the open-source ESPNet toolkit \cite{watanabe18_interspeech}. Experiments were conducted on a GeForce RTX 2080 GPU with 8GB memory. 

\section{Results and Discussion} \label{results}

Our first experiments were focused on the training setup. Therefore, several learning rates were explored until the optimal value, specified in Subsection \ref{training}, were reached. This optimum turned out to be the same for all the proposed adaptation strategies. Moreover, we studied dispensing with the scheduler, concluding that its absence not only slowed down the learning process, but also worsened the VSR system performance. 

Once determined the best setting, all the VSR systems described in Subsection \ref{method} were estimated. In this way, as reflected in Table \ref{tab:overall}, we could compare the proposed adaptation strategies in general terms. As mentioned above in Subsection \ref{method}, we explored the use of different data sets when applying our fine-tuning strategies. Thus, we were able to study how these strategies behave based on the amount of data available, since dealing with the development set means, in average terms, using an amount of data 4.5 times less than when using the training set \cite{lrec2022liprtve}. By taking into account all these aspects and the results reported in Table \ref{tab:overall}, we could infer the following conclusions:

\begin{itemize}

    \item First, regardless the fine-tuning data set used, we can observe how the MST method is significantly outperformed by the rest of the proposed strategies, a fact that supports the effectiveness of our speaker adaptation approaches.

    \item Nonetheless, when the training set was used, the MST-based system provided a considerable quality of speech recognition. This outcome could mean that the architecture employed in our experiments was able to generalise common patterns across speakers when addressing VSR.
    
    \item Regarding the amount of data used during the fine-tuning process, results reflect a drastic deterioration of system performance when the development set was used. However, this deterioration is noticeably lower when the TS-SAT strategy is applied, showing that this approach could be more robust against those situations in which a speaker presents data scarcity.
    
    \item Thus, the TS-SAT strategy stands as the best option when speaker adaptation is addressed. This fact supports the idea that a two-step fine-tuning process, where the model is first adapted to the general task, could benefit the final adaptation of the VSR system to a specific speaker.
    
    \item Finally, we consider that our results are comparable to the current state of the art in the field \cite{ma2022visual}. Moreover, our findings suggest that the fine-tuning method employed in our experiments is capable of adapting VSR end-to-end architectures in a small number of epochs, even when only a limited amount of data is available.
    
\end{itemize}

For reference, it should be noted that about 60\% WER was obtained for the speaker-independent scenario.

\begin{table}[!htbp]
    \centering
    \vspace{-0.1cm}
    \caption{System performance (WER) in average terms for each proposed adaptation strategy, depending on the data set used to fine-tune the VSR system. DEV and TRAIN refer to the development and training data set, respectively.}
    \label{tab:overall}
    \begin{tabular}{ccc}
    \toprule
    \multirow{2}{*}[-2pt]{\textbf{Strategy}} & \multicolumn{2}{c}{\textbf{Fine-Tuning Data Set}} \\
    \cmidrule(lr){2-3}
    & \footnotesize\textbf{DEV} & \footnotesize\textbf{TRAIN} \\\midrule
    \textbf{MST} & 59.6$\pm$1.3 & 36.4$\pm$1.3 \\
    \textbf{SAT} & 52.2$\pm$1.4 & 29.1$\pm$1.5 \\
    \textbf{TS-SAT} & 32.8$\pm$1.3 & 24.9$\pm$1.4
    \\\bottomrule
    \end{tabular}
    \vspace{-0.1cm}
\end{table}

On the other hand, considering only those experiments where the training set was used, we evaluated each strategy for each of the speakers selected in our study, as Figure \ref{fig:comparision} shows. From these results we could infer similar conclusions to the aforementioned ones. Nonetheless, it is noteworthy that, regardless the type of strategy applied, the VSR system provides remarkably different recognition rates depending on the speaker evaluated. Hence, a study was carried out with the aim of finding out the reasons that could explain this behaviour. For each of these speakers, several statistics were computed, such as the number of words per utterance, the perplexity of the LM in the test samples, or the number of training seconds. Thus, it was analyzed how, as the word error rate increased, each of the statistics varied. Nonetheless, we were not able to identify any trends or patterns from the data. Therefore, we could say that these experiments suggested that the reason why VSR systems behaved in this way could be related to aspects that are difficult to model, such as better vocalizations or certain oral physiognomies that reflect more adequately speech articulations.
\vspace{-0.15cm}
\section{Conclusions and Future Work}

In our research, the continuous VSR for the Spanish language has been addressed from a speaker-adapted perspective. The challenging LIP-RTVE database was used throughout all our experiments. The VSR systems defined in our work were based on a hybrid CTC/Attention architecture \cite{ma2022visual}. Thus, it was studied how the estimation of specialized end-to-end systems for a specific speaker could affect the quality of speech recognition. Hence, different speaker adaptation strategies based on the fine-tuning technique were proposed. Our findings showed that a two-step fine-tuning process, where the VSR system is first adapted to the task domain, provided significant improvements when the speaker adaptation was addressed. Furthermore, results comparable to the current state of the art \cite{ma2022visual} were reached even when only a limited amount of data was available.

Regarding our future work, we consider exploring more sophisticated approaches as those described in \cite{weninger19b_interspeech,ctc2018adaption,delcroix18_interspeech,watanabe2018adaptation} when more powerful GPUs are available. Additionally, we propose to study how using auxiliary tasks, as proposed in \cite{ma2022visual}, could benefit the quality of speech recognition. The use of distillation-based learning methods \cite{ji2021distill,afouras2020distilling} to estimate a simpler model with similar performance is another line to follow. Besides, we consider the integration of a previous speaker identification module \cite{facerecognition2018}. Thus, unlike our experiments where a perfect speaker classifier was assumed, we will be able to develop a system aimed at a realistic application.


\vspace{-0.15cm}
\section{Acknowledgements}

This work was partially supported by Generalitat Valenciana under project DeepPattern (PROMETEO/219/121) and by Grant PID2021-124719OB-I00 funded by MCIN/AEI/10.13039/501100011033/ ERDF, EU.

\bibliographystyle{IEEEtran}

\bibliography{main}

\begin{thebibliography}{10}
\providecommand{\url}[1]{#1}
\csname url@samestyle\endcsname
\providecommand{\newblock}{\relax}
\providecommand{\bibinfo}[2]{#2}
\providecommand{\BIBentrySTDinterwordspacing}{\spaceskip=0pt\relax}
\providecommand{\BIBentryALTinterwordstretchfactor}{4}
\providecommand{\BIBentryALTinterwordspacing}{\spaceskip=\fontdimen2\font plus
\BIBentryALTinterwordstretchfactor\fontdimen3\font minus
  \fontdimen4\font\relax}
\providecommand{\BIBforeignlanguage}[2]{{%
\expandafter\ifx\csname l@#1\endcsname\relax
\typeout{** WARNING: IEEEtran.bst: No hyphenation pattern has been}%
\typeout{** loaded for the language `#1'. Using the pattern for}%
\typeout{** the default language instead.}%
\else
\language=\csname l@#1\endcsname
\fi
#2}}
\providecommand{\BIBdecl}{\relax}
\BIBdecl

\bibitem{mcgurk1976hearing}
H.~McGurk and J.~MacDonald, ``Hearing lips and seeing voices,'' \emph{Nature},
  vol. 264, no. 5588, pp. 746--748, 1976.

\bibitem{juang1991adverse}
B.~Juang, ``Speech recognition in adverse environments,'' \emph{Computer Speech
  \& Language}, vol.~5, no.~3, pp. 275--294, 1991.

\bibitem{afouras2018deep}
T.~Afouras, J.~S. Chung, A.~Senior, O.~Vinyals, and A.~Zisserman, ``Deep
  audio-visual speech recognition,'' \emph{IEEE Transactions on Pattern
  Analysis and Machine Intelligence}, 2018.

\bibitem{ma2022visual}
P.~Ma, S.~Petridis, and M.~Pantic, ``Visual speech recognition for multiple
  languages in the wild,'' \emph{arXiv preprint arXiv:2202.13084}, 2022.

\bibitem{fernandez2018survey}
A.~Fernandez-Lopez and F.~M. Sukno, ``Survey on automatic lip-reading in the
  era of deep learning,'' \emph{Image and Vision Computing}, vol.~78, pp.
  53--72, 2018.

\bibitem{afouras2021sub}
K.~R. Prajwal, T.~Afouras, and A.~Zisserman, ``Sub-word level lip reading with
  visual attention,'' \emph{arXiv preprint arXiv:2110.07603}, 2021.

\bibitem{duchnowski2000development}
P.~Duchnowski, D.~S. Lum, J.~C. Krause, M.~G. Sexton, M.~S. Bratakos, and L.~D.
  Braida, ``Development of speechreading supplements based on automatic speech
  recognition,'' \emph{IEEE trans. on biomedical engineering}, vol.~47, no.~4,
  pp. 487--496, 2000.

\bibitem{leung04_interspeech}
K.-Y. Leung, M.-W. Mak, and S.-Y. Kung, ``{Articulatory feature-based
  conditional pronunciation modeling for speaker verification},'' in
  \emph{Proc. Interspeech}, 2004, pp. 2597--2600.

\bibitem{cox2008challenge}
S.~J. Cox, R.~W. Harvey, Y.~Lan, J.~L. Newman, and B.-J. Theobald, ``The
  challenge of multispeaker lip-reading.'' in \emph{AVSP}, 2008, pp. 179--184.

\bibitem{thangthai2017}
K.~Thangthai and R.~Harvey, ``Improving computer lipreading via dnn sequence
  discriminative training techniques,'' in \emph{Proc. Interspeech}, 2017, pp.
  3657--3661.

\bibitem{assael2016lipnet}
Y.~M. Assael, B.~Shillingford, S.~Whiteson, and N.~de~Freitas, ``Lipnet:
  Sentence-level lipreading,'' \emph{ArXiv}, vol. abs/1611.01599, 2016.

\bibitem{watanabe2018adaptation}
T.~Ochiai, S.~Watanabe, S.~Katagiri, T.~Hori, and J.~Hershey, ``Speaker
  adaptation for multichannel end-to-end speech recognition,'' in
  \emph{ICASSP}, 2018, pp. 6707--6711.

\bibitem{delcroix18_interspeech}
M.~Delcroix, S.~Watanabe, A.~Ogawa, S.~Karita, and T.~Nakatani, ``{Auxiliary
  Feature Based Adaptation of End-to-end ASR Systems},'' in \emph{Proc.
  Interspeech}, 2018, pp. 2444--2448.

\bibitem{weninger19b_interspeech}
F.~Weninger, J.~Andrés-Ferrer, X.~Li, and P.~Zhan, ``{Listen, Attend, Spell
  and Adapt: Speaker Adapted Sequence-to-Sequence ASR},'' in \emph{Proc.
  Interspeech 2019}, 2019, pp. 3805--3809.

\bibitem{ctc2018adaption}
K.~Li, J.~Li, Y.~Zhao, K.~Kumar, and Y.~Gong, ``Speaker adaptation for
  end-to-end ctc models,'' in \emph{IEEE SLT}, 2018, pp. 542--549.

\bibitem{kandala19_interspeech}
P.~A. Kandala, A.~Thanda, D.~K. Margam, R.~C. Aralikatti, T.~Sharma, S.~Roy,
  and S.~M. Venkatesan, ``{Speaker Adaptation for Lip-Reading Using Visual
  Identity Vectors},'' in \emph{Proc. Interspeech}, 2019, pp. 2758--2762.

\bibitem{cogans2020adriana}
A.~Fernandez-Lopez, A.~Karaali, N.~Harte, and F.~M. Sukno, ``Cogans for
  unsupervised visual speech adaptation to new speakers,'' in \emph{ICASSP
  2020}, 2020, pp. 6294--6298.

\bibitem{ssi2020review}
J.~A. Gonzalez-Lopez, A.~Gomez-Alanis, J.~M. Martín~Doñas, J.~L.
  Pérez-Córdoba, and A.~M. Gomez, ``Silent speech interfaces for speech
  restoration: A review,'' \emph{IEEE Access}, vol.~8, pp. 177\,995--178\,021,
  2020.

\bibitem{kenji2020enhancement}
K.~Matsui, K.~Fukuyama, Y.~Nakatoh, and Y.~O. Kato, ``Speech enhancement system
  using lip-reading,'' in \emph{IEEE 2nd IICAIET}, 2020, pp. 1--5.

\bibitem{fernandez2017towards}
A.~Fernandez-Lopez, O.~Martinez, and F.~M. Sukno, ``Towards estimating the
  upper bound of visual-speech recognition: The visual lip-reading feasibility
  database,'' in \emph{12th FG}.\hskip 1em plus 0.5em minus 0.4em\relax IEEE,
  2017, pp. 208--215.

\bibitem{gimenogomez21_iberspeech}
D.~Gimeno-Gómez and C.-D. Martínez-Hinarejos, ``{Analysis of Visual Features
  for Continuous Lipreading in Spanish},'' in \emph{Proc. IberSPEECH}, 2021,
  pp. 220--224.

\bibitem{lrec2022liprtve}
------, ``L{IP-RTVE}: {A}n {A}udiovisual {D}atabase for {C}ontinuous {S}panish
  in the {W}ild,'' in \emph{Proceedings of LREC}.\hskip 1em plus 0.5em minus
  0.4em\relax European Language Resources Association, June 2022, pp.
  2750--2758.

\bibitem{afouras2018lrs3}
T.~Afouras, J.~Chung, and A.~Zisserman, ``Lrs3-ted: a large-scale dataset for
  visual speech recognition,'' \emph{arXiv preprint arXiv:1809.00496}, 2018.

\bibitem{gales2008application}
M.~Gales and S.~Young, \emph{The application of hidden Markov models in speech
  recognition}.\hskip 1em plus 0.5em minus 0.4em\relax Now Publishers Inc,
  2008.

\bibitem{ivectors2013}
G.~Saon, H.~Soltau, D.~Nahamoo, and M.~Picheny, ``Speaker adaptation of neural
  network acoustic models using i-vectors,'' in \emph{IEEE ASRU}, 2013, pp.
  55--59.

\bibitem{graves2006ctc}
A.~Graves, S.~Fern\'{a}ndez, F.~Gomez, and J.~Schmidhuber, ``Connectionist
  temporal classification: Labelling unsegmented sequence data with recurrent
  neural networks,'' in \emph{Proceedings of the 23rd ICML}.\hskip 1em plus
  0.5em minus 0.4em\relax ACM, 2006, p. 369–376.

\bibitem{zadeh2020moseas}
A.~B. Zadeh, Y.~Cao, S.~Hessner, P.~P. Liang, S.~Poria, and L.-P. Morency,
  ``C{MU-MOSEAS}: A multimodal language dataset for spanish, portuguese, german
  and french,'' in \emph{Proceedings of EMNLP}, 2020, pp. 1801--1812.

\bibitem{adriana2022alr}
A.~Fernandez-Lopez and F.~M. Sukno, ``End-to-end lip-reading without
  large-scale data,'' \emph{IEEE/ACM Transactions on Audio, Speech, and
  Language Processing}, vol.~30, pp. 2076--2090, 2022.

\bibitem{he2016resnet}
K.~He, X.~Zhang, S.~Ren, and J.~Sun, ``Deep residual learning for image
  recognition,'' in \emph{IEEE CVPR}, 2016, pp. 770--778.

\bibitem{gulati20_interspeech}
A.~Gulati, J.~Qin, C.-C. Chiu, N.~Parmar, Y.~Zhang, J.~Yu, W.~Han, S.~Wang,
  Z.~Zhang, Y.~Wu, and R.~Pang, ``{Conformer: Convolution-augmented Transformer
  for Speech Recognition},'' in \emph{Proc. Interspeech}, 2020, pp. 5036--5040.

\bibitem{vaswani2017attention}
A.~Vaswani, N.~Shazeer, N.~Parmar, J.~Uszkoreit, L.~Jones, A.~N. Gomez,
  {\L}.~Kaiser, and I.~Polosukhin, ``Attention is all you need,''
  \emph{NeurIPS}, vol.~30, 2017.

\bibitem{watanabe2017ctcattention}
S.~Watanabe, T.~Hori, S.~Kim, J.~R. Hershey, and T.~Hayashi, ``Hybrid
  ctc/attention architecture for end-to-end speech recognition,'' \emph{IEEE
  Journal of Selected Topics in Signal Processing}, vol.~11, no.~8, pp.
  1240--1253, 2017.

\bibitem{petridis2018ctc-attn}
S.~Petridis, T.~Stafylakis, P.~Ma, G.~Tzimiropoulos, and M.~Pantic,
  ``Audio-visual speech recognition with a hybrid ctc/attention architecture,''
  in \emph{IEEE SLT}, 2018, pp. 513--520.

\bibitem{deng2020retina}
J.~Deng, J.~Guo, E.~Ververas, I.~Kotsia, and S.~Zafeiriou, ``Retinaface:
  Single-shot multi-level face localisation in the wild,'' in \emph{IEEE/CVF
  CVPR}, 2020, pp. 5202--5211.

\bibitem{bulat2017facealign}
A.~Bulat and G.~Tzimiropoulos, ``How far are we from solving the 2d \& 3d face
  alignment problem? (and a dataset of 230,000 3d facial landmarks),'' in
  \emph{IEEE ICCV}, 2017, pp. 1021--1030.

\bibitem{zhang2020rois}
Y.~Zhang, S.~Yang, J.~Xiao, S.~Shan, and X.~Chen, ``Can we read speech beyond
  the lips? rethinking roi selection for deep visual speech recognition,'' in
  \emph{15th IEEE FG}, 2020, pp. 356--363.

\bibitem{salesky21_interspeech}
E.~Salesky, M.~Wiesner, J.~Bremerman, R.~Cattoni, M.~Negri, M.~Turchi, D.~W.
  Oard, and M.~Post, ``{The Multilingual TEDx Corpus for Speech Recognition and
  Translation},'' in \emph{Proc. Interspeech}, 2021, pp. 3655--3659.

\bibitem{bisani2004bootstrap}
M.~Bisani and H.~Ney, ``Bootstrap estimates for confidence intervals in asr
  performance evaluation,'' in \emph{ICASSP}, vol.~1.\hskip 1em plus 0.5em
  minus 0.4em\relax IEEE, 2004, pp. 409--412.

\bibitem{watanabe18_interspeech}
S.~Watanabe, T.~Hori, S.~Karita, T.~Hayashi, J.~Nishitoba, Y.~Unno, N.~{Enrique
  Yalta Soplin}, J.~Heymann, M.~Wiesner, N.~Chen, A.~Renduchintala, and
  T.~Ochiai, ``{ESPnet: End-to-End Speech Processing Toolkit},'' in \emph{Proc.
  Interspeech}, 2018, pp. 2207--2211.

\bibitem{ji2021distill}
J.~W. Yoon, H.~Lee, H.~Y. Kim, W.~I. Cho, and N.~S. Kim, ``Tutornet: Towards
  flexible knowledge distillation for end-to-end speech recognition,''
  \emph{IEEE/ACM Transactions on Audio, Speech, and Language Processing},
  vol.~29, pp. 1626--1638, 2021.

\bibitem{afouras2020distilling}
T.~Afouras, J.~S. Chung, and A.~Zisserman, ``A{SR} is all you need: Cross-modal
  distillation for lip reading,'' in \emph{ICASSP}, 2020, pp. 2143--2147.

\bibitem{facerecognition2018}
M.~Uzun-Per and M.~Gökmen, ``Face recognition with patch-based local walsh
  transform,'' \emph{Signal Processing: Image Communication}, vol.~61, pp.
  85--96, 2018.

\end{thebibliography}

\end{document}